%% file: references.tex
\documentclass[runningheads]{llncs}
\usepackage{graphicx}
\usepackage{amsmath}
\usepackage{listings}
\usepackage{wrapfig}
\usepackage{hyperref}

\begin{document}
\title{Towards Complex Ontology Alignment using Large Language Models}
%
%
\author{Reihaneh Amini\inst{1}\and
Sanaz Saki Norouzi\inst{1}\and
Pascal Hitzler\inst{1}\and
Reza Amini \inst{2}}
\authorrunning{R. Amini, S. Norouzi, P. Hitzler, R. Amini}
%
\institute{Kansas State University, Manhattan, KS, USA \email{\{reihanea,sanazsn,hitzler\}@ksu.edu} \\
\and Wright State University, Dayton, OH, USA \email{amini.4@wright.edu}}

\maketitle              
\begin{abstract} Ontology alignment, a critical process in the Semantic Web for detecting relationships between different ontologies, has traditionally focused on identifying so-called ``simple'' 1-to-1 relationships through class labels and properties comparison. The more practically useful exploration of more complex alignments remains a hard problem to automate, and as such is largely underexplored, i.e. in application practice it is usually done manually by ontology and domain experts.
Recently, the surge in Natural Language Processing (NLP) capabilities, driven by advancements in Large Language Models (LLMs), presents new opportunities for enhancing ontology engineering practices, including ontology alignment tasks. This paper investigates the application of LLM technologies to tackle the complex ontology alignment challenge. Leveraging a prompt-based approach and integrating rich ontology content -- so-called \emph{modules} -- our work constitutes a significant advance towards automating the complex alignment task.

\end{abstract}
\input{1-introduction}


\section{Related Work}\label{sec:related}
In the semantic web domain, ontology alignment is crucial to discerning correspondences between concepts and properties across various ontologies. It involves two primary types of matching: simple and complex alignment \cite{shvaiko2011ontology}. Simple alignment focuses on identifying 1-to-1 equivalence relationships between concepts in two different ontologies. Complex alignment aims to match 1-to-n, n-to-1, or m-to-n equivalence or subsumption relationships between complex expressions in two ontologies (see e.g. \cite{zhou2020geolink}); the alignments are usually expressed in form of (first-order logic) rules, description logic axioms, or similar logical or quasi-logical notation. 

Most of the work on ontology alignment is evaluated by the OAEI, an annual event focusing on ontology-matching systems’ performance, starting in 2004, with focus mostly on simple alignment. A complex alignment track was started in 2018 because simple matching is often not sufficient to capture the rich semantics needed for advanced applications using realistic ontologies. In the context of this track, the datasets utilized throughout these years encompass Complex Conference, Populated Complex Conference, Hydrography, GeoLink, Populated GeoLink, Populated Enslaved, and Taxon. It should be noted, however, that some datasets have ceased to be evaluated in recent years \cite{algergawy2018results,zhou2020geolink,DBLP:conf/om2/PourABCCCC0FFFH23} as each track depends on volunteers to run it. 

The GeoLink complex alignment benchmark stands out as a real-world data\-set for testing complex ontology alignment; using real-world datasets like GeoLink for testing tasks such as ontology alignment is crucial as it reflects the complexities of practical applications, helps validate the robustness and versatility of ontology alignment algorithms, and ensures their effectiveness beyond synthetic ones. It contains two ontologies, the GeoLink Base Ontology (GBO) and the Geo\-Link Modular Ontology (GMO). These are paired with a reference alignment created with help from domain specialists \cite{zhou2020geolink,amini2020geolink}. During these years, only two or three systems registered for evaluation in this track. Evaluation results show \cite{DBLP:conf/semweb/AlgergawyFFFHHJ19,DBLP:conf/semweb/PourAAFFHHJJKKL20,DBLP:conf/semweb/PourAAAFFFHHHHI21} that the systems that make use of instance data that is shared between the two ontologies perform relatively well, with high precision but low recall, i.e. many matches are missed. Furthermore, the availability of shared instance data, as for this benchmarking competition, while helpful for advancing the state of the art, is unrealistic for most practical application scenarios for ontology alignment: Usually, such shared data would not be availble. The new results which we present in this paper have been obtained without taking shared instance data into consideration, which, in our opinion, constitutes a major advance over the state of the art.

Currently, there is a plethora of reserach on applications of LLMs in a wide variety of fields, often with significant success. In the field of ontology alignment, \cite{DBLP:conf/om2/NorouziMH23} presents a zero-shot evaluation of ChatGPT-4 using the OAEI conference track ontologies (consisting of small and medium sized ontologies), showing a high recall and F1 of 0.52. Also, \cite{DBLP:conf/semweb/0008C0023} provides a zero-shot evaluation of GPT 3.5 and Flan-T5-XXL on the OAEI Bio-ML track, and they also achieved low precision and high recall. \cite{hertling2023olala} employs a comparable approach implemented in MELT, and the candidate generation and matching processes work in collaboration with Sentence-BERT, their evaluation has been done on some OAEI track datasets which almost achieve high F1 scores. In \cite{DBLP:journals/corr/abs-2312-00326}, they present a methodology utilizing LLM-based agents for retrieval and matching processes, which they enhance by incorporating Retrieval Augmented Generation (RAG) within their agents. The assessment was conducted on several tracks of OAEI. The results from the conference track indicate a high recall but low precision. The performance of the proposed method on other tracks is relatively good. However all these results do not go beyond simple alignment. In particular, no good results for complex alignment benchmarks have been reported yet.

 
\bigskip
 

\input{3-method}

\input{4-eval}

\input{5-discussion}

\section{Conclusion}

We have presented the very first approach that is able to achieve good accuracy for complex ontology alignment without relying on shared individuals. The system is neural-symbolic in its nature as it addresses a symbolic task (complex ontology alignment), the output of which are alignment rules expressed in some logic, it furthermore makes decisive use of additional symbolic input in the form of ontology modules, and it uses an LLM as core processing engine. Our results suggest that further work bear the promise to result in strong complex ontology alignment systems, for ontologies that carry sufficient internal structure. 


\paragraph{Acknowledgments} We acknowledge partial support by NSF award no. 2333532 "Proto-OKN Theme 3: An Education Gateway for the Proto-OKN".

\bibliographystyle{plain}
\bibliography{references}


\end{document}

%% file: 1-introduction.tex
\section{Introduction}

Ontology alignment (sometimes called ontology matching) \cite{DBLP:books/daglib/0032976} is the task of establishing mappings between different ontologies, and as a research field it is concerned with ways to automate or at least semi-automate this task. For those not familiar with the field: Ontologies, which are usually knowledge bases expressed using Description Logics \cite{FOST} (including the W3C Web Ontology Language -- OWL -- standard \cite{owl-primer}) in this case act as a type of data schema for data expressed as knowledge graphs \cite{DBLP:journals/cacm/Hitzler21}, i.e. the establishing of mappings between ontologies is central for schema-based data integration purposes. 

Ontology alignment has been studied for over two decades, resulting in the development of many alignment approaches and systems. The majority of these systems are designed to detect only so-called ``simple'' 1-to-1 mappings between ontologies, primarily by establishing equivalence relationships between classes (unary predicates), or between properties (binary relationships); for example, one ontology may have a class called ``Person'' while another may have a class called ``Human'', and an ontology alignment mapping may state that these two classes are in fact equivalent. It has long been recognized in the Semantic Web and Ontologies community that such simple mappings are helpful but ultimately insufficient for data integration task, for which mappings would need to be in the form of complex mapping rules\footnote{See Section \ref{sec:eval} for an example.}, that can be expressed, e.g., as Datalog rules. However, detecting complex alignments between ontologies remains a very challenging and thus largely unexplored area with only few contributions that made progress in restricted settings (see Section \ref{sec:related}). In current practice, establishing complex alignments between two or more ontologies requires domain experts to collaborate and manually generate the alignments, and this is usually a very work-intensive and thus expensive task. Any automation or semi-automation would have significant added value.

The Ontology Alignment Evaluation Initiative (OAEI)\footnote{http://oaei.ontologymatching.org/} is a long-standing coordinated international effort aimed at improving and evaluating ontology alignment and coreference resolution technologies.\footnote{Co-reference resolution is about establishing equivalence and non-equivalence mappings between individuals, or constants, a related but different task of similar practical importance.} It organizes annual evaluation campaigns \cite{DBLP:conf/om2/PourABCCCC0FFFH23} that provide a controlled environment where participants can test their ontology alignment systems using various benchmark tests. The benchmarks cover a range of complexity levels and real-world scenarios, aiming to simulate different aspects of the ontology alignment process.

With significant advancements in the natural language processing (NLP) and natural language understanding (NLU) fields, spurred by Large Language Models (LLMs), it has become possible to extract meanings from text and reason about it more effectively. OpenAI\footnote{https://openai.com} has been at the forefront of this research, developing the Generative Pre-trained Transformer (GPT) series of models, which have attracted considerable attention from researchers, developers, and users. One of the most notable models, introduced in March 2022, was GPT-4 \cite{achiam2023gpt}. This transformer-based model is designed to predict the next token in generating text and has shown improvements in producing results that more closely align with user intent compared to its predecessor, GPT-3.5, on 70.2\% of the prompts.\footnote{See OpenAI: Introducing ChatGPT, 2022, https://openai.com/blog/chatgpt and Greg Brockman, Peter Welinder, Mira Murati, and OpenAI, 2020, https://openai.com/blog/openai-api.}

Recent advancements in applying Large Language Models (LLMs) to Semantic Web and ontology engineering tasks have shown promising results, due to the importance of NLP for such tasks. A notable study  \cite{hertling2023olala} demonstrated the effectiveness of zero-shot and few-shot prompting with LLMs on various tasks within the Ontology Alignment Evaluation Initiative (OAEI), highlighting their potential in this area. The study was restricted to simple alignments. Indeed, as we will see later, a straightforward tasking of LLMs with the production of complex alignments does not quite work.

One of the difficulties with ontology alignment is that ontologies often tend to be underspecified and with little internal structure that may add some self-explainability. This can be seen for example by the considerable disagreement between humans as to ``correct'' alignments, even for the simple alignment task \cite{cheatham2014conference}. It has been posited that additional internal structure, e.g. in the form of conceptual "ontology modules" should aid with ontology engineering tasks that are hard to automate \cite{shimizu2023modular}. Following this hitherto merely conceptual argument, we made use of ontology modules in our approach to generate complex alignments, and as we will report below, our prompt-based approach for discovering complex alignments between ontologies yields significantly better results when richer content in the form of ontology modules is available.

Since ontologies are knowledge bases expressed using formal logic, and mapping rules are also expressed using formal logic and processed as such, ontology alignment is a key symbolic task that we are here addressing usine "neural" means (i.e. LLMs as artificial neural networks), as such contributing to the body of approaches and methods for neural-symbolic learning and reasoning \cite{DBLP:series/faia/369}.

The structure of this paper is as follows: Section \ref{sec:related} reviews past research and the current state-of-the-art methods in complex alignment. Section \ref{sec:method} outlines the methodologies and approaches we utilized to experiment with an effective prompt-based method for discovering complex alignments. The evaluation of our approach and the specifics of prompt tailoring are discussed in Sections \ref{sec:eval} and \ref{sec:discussion}, respectively.

%% file: 3-method.tex
\section{Complex Alignment by Large Language Model}\label{sec:method}

\subsection{Ontology Modules}
\label{method-1}

Ontologies that are conceptually clear are inherently more reusable, primarily because they are straightforward to comprehend from the outset. Consequently, a primary objective in the field of ontology research is to develop ontology modeling methodologies for creating ontologies with high conceptual clarity. It is a fundamental assumption that adhering to well-established modeling principles enhances the understandability and reusability of an ontology \cite{shimizu2023modular}.

It is known that repurposing or adjusting of ontologies for new applications often presents challenges. These challenges stem from various factors, including: (i) mismatches in the level of detail between ontologies and the intended use-cases, (ii) unclear concepts within the ontologies intended for reuse, (iii) the absence and the complexity of ontology alignment practices, and (iv) the scarcity of tools designed to facilitate ontology reuse and support the process. To tackle these issues, the Modular Ontology Modeling (MOMo) approach was proposed \cite{shimizu2023modular}.
In the MOMo framework, a key feature is the creation and connection of compact, independent modules, which offers numerous advantages. For instance, it streamlines maintenance because changes to a single module have little to no effect on the entire ontology. Additionally, tracking the origin of every ontology segment back to the initial requirements becomes straightforward through the use of module documentation or metadata.

\begin{figure}[tb]
\centering
 \includegraphics[width=0.8\textwidth]{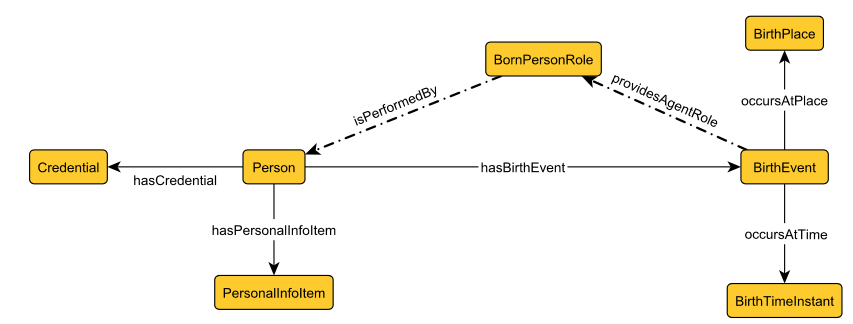}
 \caption{A diagram for a Person module} \label{fig-pr-5}
\end{figure}

In the Semantic Web community, the term \textbf{``module''} can mean many things. In our case, it refers to a specific part of an ontology that encapsulates a principal concept and its main features, such as a Person module capturing details about ``BirthEvent'', ``PersonalInfoItem'', and ``Credential'' as depicted in Figure \ref{fig-pr-5}.  Modules serve dual roles: they are technical constructs that, on the one hand, demarcate parts of ontologies that group related classes and their interactions and, on the other hand, they do this in a way that aligns with domain experts' understanding. Despite potential overlaps and hierarchical structures within them, modules organize the ontology into a network of interrelated pieces, each mirroring the domain's conceptual framework as understood by experts. 

Modules enable a strategic approach to ontology modeling by allowing the work to be broken down into manageable segments; initially focusing on individual modules before linking them together. This approach provides a clear way to manage the complexities of large, cohesive ontologies by breaking down the process into understanding individual modules and their interconnections. This modular approach also aligns with the way domain experts conceptualize their fields, making both the ontology and its documentation more accessible and comprehensible. Since each module can be easily swapped out for another—perhaps one that offers a different level of detail—modifications are localized, making the entire system more adaptable \cite{krisnadhi2015geolink}. 

In the research we present herein, we explore the impact of integrating module information on the effectiveness of identifying complex alignments. Specifically, we include descriptions, Core Axioms (where applicable), and Alignment information as outlined in the GeoLink Core Ontology Design Patterns \cite{krisnadhi2015geolinkcorepattern} that constitute the GeoLink Modular Ontology (GMO), that had been developed as an integrated schema for combining several large-scale ocean science data repositories \cite{cheatham2018geolink}. As we will see further below, the utilization of additional module information is of core importance in order to solve the complex alignment problem, in our setting. 

\subsection{Design of the Prompting Process}

Tailoring an LLM for specific tasks can be achieved either by fine-tuning the model with select data or by using prompts for in-context learning. Fine-tuning requires significant computational resources and expertise, which may not be feasible for all users or organizations. Moreover, it risks over-fitting the model to the training data, potentially diminishing its ability to generalize to other tasks. In contrast, employing various prompting techniques on a fixed LLM is more resource-efficient, requiring less computational power, time, and expertise. Additionally, prompts allow for the flexible and on-the-fly adaptation of the model to a wide range of tasks by applying different prompting techniques \cite{liu2023pre}. 

Prompt engineering is increasingly vital for enhancing the performance of large language models (LLMs) across a wide range of tasks. These models have shown impressive capabilities using zero-shot prompts, where they generate responses without prior specific training examples. However, for more complex tasks requiring deeper reasoning, advanced prompting techniques are necessary. Incorporating context into prompts significantly improves model performance. Techniques such as few-shot prompting, where the model is given a few examples to learn from, or chain-of-thought prompting, which guides the model through a series of reasoning steps, are particularly effective in enhancing the model's understanding and response accuracy \cite{white2023prompt}. 

There is a range of task-agnostic prompting techniques available for use with large language models.\footnote{https://www.promptingguide.ai/techniques} These include:
\begin{itemize}
    \item Zero-shot prompting: This technique involves providing a single, natural language description of the task at inference time, without any prior examples.
    \item Few-shot prompting: This approach includes giving the model a few examples of the task, complete with context and successful outcomes, to guide its understanding and responses \cite{brown2020language}.
    \item Chain-of-thought prompting: This method involves supplying the model with a series of thought process examples, helping it to navigate through reasoning steps to arrive at a conclusion \cite{wei2022chain}. 
\end{itemize}

These techniques enable large language models to adapt and respond to a wide variety of tasks with varying levels of guidance and specificity. 

\begin{wrapfigure}[13]{r}{36mm}
\vspace*{-10mm}
 \includegraphics[keepaspectratio,width=30mm]{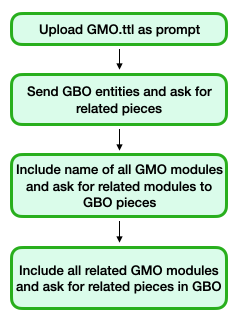}
 \caption{Prompting workflow} \label{pr_workflow}
\end{wrapfigure}

Figure \ref{pr_workflow} illustrates our workflow, which begins by uploading the entire GMO file as an initial prompt. Subsequently, we include specific entities from the second ontology, the GBO, to inquire about their alignment with the GMO. If the initial prompt successfully provides the relevant segments, further prompts in the chain-of-thought process are unnecessary. Otherwise, we can provide a list of all GMO module names and ask GPT to identify the most related modules for the GBO entities. Then, in a subsequent prompt,  we supplement the inquiry with module information and request the related segments again.

In the following section, we examine the performance of GPT-4 in response to various entities, modules, and prompts it processes.

%% file: 4-eval.tex
\section{Evaluation}\label{sec:eval}

For the evaluation of our methodology, we employ the GeoLink Complex Alignment dataset previously detailed in Zhou et al., 2018 \cite{zhou2018complex}. This dataset comprises 109 complex alignment rules between GMO and GBO, each of which has been thoroughly analyzed to test our hypothesis. Detailed findings from this study, including both the prompts used and the responses obtained, are available online.\footnote{https://daselab.cs.ksu.edu/publications/alignment-rules-gbo-gmo}

\subsection{Prompting and Detailed Example}

\begin{figure}[tb]
\centering
 \includegraphics[keepaspectratio,height=7cm, width=10cm]{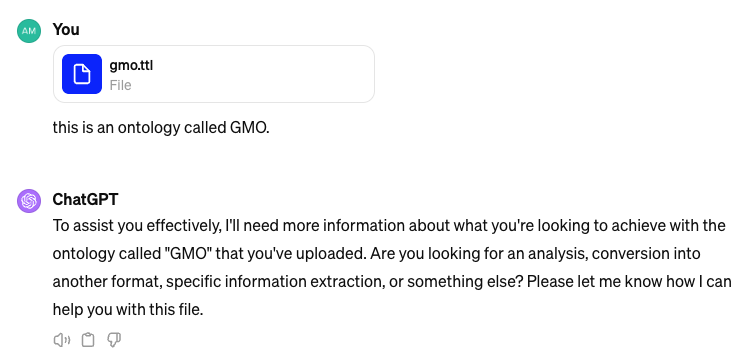}
 \caption{Uploading GMO.ttl file as prompt.} \label{FIG-pr-gmo-ttl}
\end{figure}

OpenAI's functionality includes the ability to load a prompt from a file, triggering backend processing to parse the uploaded data. We uploaded the entire GMO ontology RDF file in turtle (.ttl) format. OpenAI GPT-4 parsed the data, albeit with some latency, prompting a request for more specific tasks to be directed to it (Figure \ref{FIG-pr-gmo-ttl}). 

The complex alignment rule we are targeting in this example is

\begin{align}
&\mathtt{Award}(x) \land \mathtt{hasCoPrincipalInvestigator}(x,z) \leftrightarrow \nonumber \\
&\mathtt{FundingAward}(x) \land \mathtt{providesAgentRole}(x,y) \label{eq:example-rule}\\
&\land \mathtt{CoPrincipalInvestigatorRole}(y) \land \mathtt{performedBy}(y,z) \nonumber
\end{align}

where GBO entities $\mathtt{has\-CoPrincipal\-In\-vestigator}$ (a relation or binary predicate or so-called object property)  and $\mathtt{Award}$ (a class or unary predicate) can be found on the left-hand side, and the GMO entities on the right-hand side. This is an example of a complex alignment rule expressed in first-order predicate logic and our objective is to assess the effectiveness of detecting complex alignments by prompting GPT-4. 

After uploading the GMO, we extract the GBO entities found on the left-hand side of the complex alignment rule (i.e., \textbf{“Award”} and \textbf{“hasCoPrincipalInvestigator”}) from the GBO RDF file in its original format. For our example, this looks as follows. 

\makeatletter
\def\verbatim@font{\normalfont\ttfamily\small}
\makeatother

\begin{verbatim}
###  http://gbo#Award
main:Award rdf:type owl:Class ;
           rdfs:comment "Funding provided by an Organization
           enabling Participation.";
           rdfs:label "Award" . 
###  http://gbo#hasCoPrincipalInvestigator
main:hasCoPrincipalInvestigator rdf:type owl:ObjectProperty ;
                    owl:inverseOf main:isCoPrincipalInvestigatorOf ;
                    rdfs:domain [ rdf:type owl:Class ;
                                  owl:unionOf ( main:Award
                                                main:Program
                                              )
                                ] ;
                    rdfs:range main:Person ;
                    rdfs:label "hasCoPrincipalInvestigator" .
\end{verbatim}

\begin{figure}[tb]
\centering
 \includegraphics[keepaspectratio,height=6cm, width=10cm]{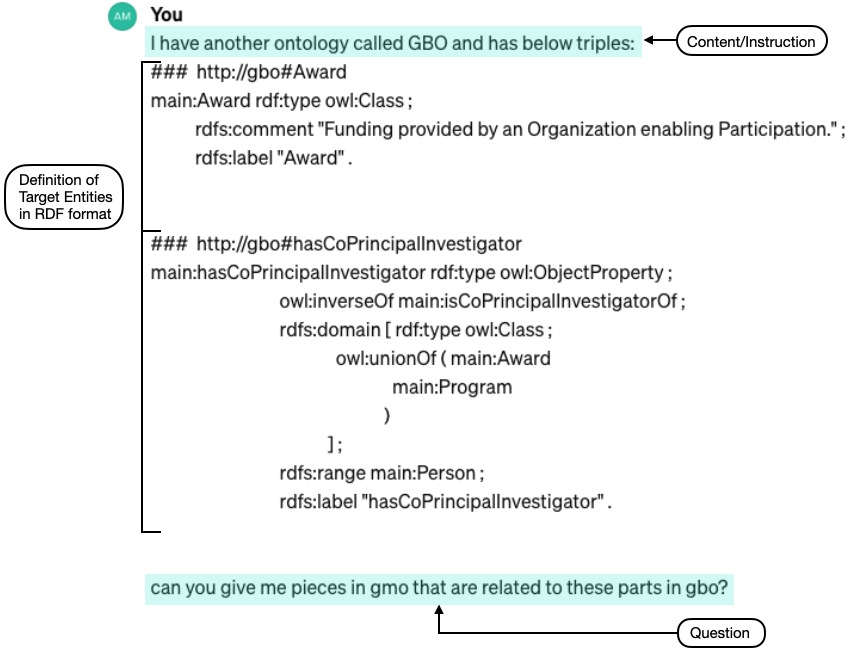}
 \caption{GBO-related instruction and question in prompt} \label{fig-pr-2}
\end{figure}

Next, our prompt instructs GPT-4 to examine the components in GMO that are associated with these elements in GBO, as depicted in Figure \ref{fig-pr-2}. 


What we observed from GPT-4's responses in this step is often not a comprehensive answer. Many times, it explicitly states that \emph{It appears there were no results found in the ontology file ``gmo.ttl'' directly related to the terms you mentioned from the "GBO" ontology, such as ``Award'', ``hasCoPrincipalInvestigator'', ``Person'', ``Program'', or ``isCoPrincipalInvestigatorOf''.} GPT-4 typically continues its response with further inquiries such as \emph{To proceed, I can manually examine the ``gmo.ttl'' file to identify any related concepts or properties that might align with those in the ``GBO'' ontology. This process involves looking for classes or properties that serve similar purposes or are conceptually related, even if they are not directly named the same. Would you like me to perform this manual examination, or is there another way I can assist you with these ontologies?}

\begin{figure}[tb]
\centering
 \includegraphics[keepaspectratio, width=.8\textwidth]{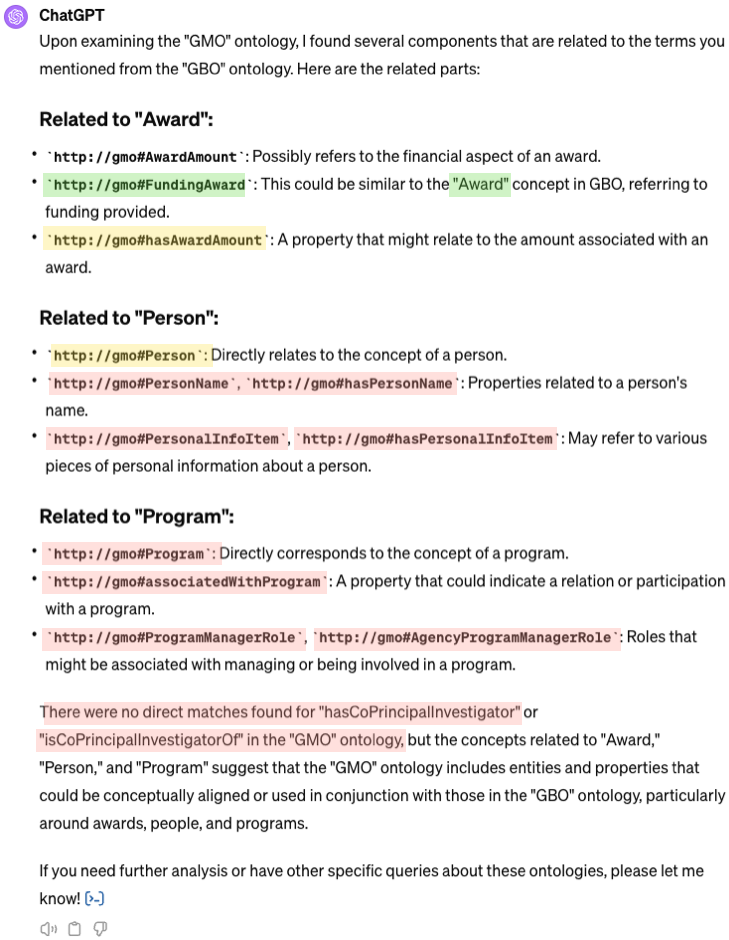}
 \caption{GPT-4 response to our initial question} \label{fig-pr-3}
\end{figure}

Our consistent response to this question has always been ``Yes'', and the typical output we received from GPT is illustrated in Fig \ref{fig-pr-3}. As depicted, the LLM successfully identifies a similar entity to \textbf{gbo\#Award}, which in GMO is \textbf{gmo\#AwardAmount} (highlighted in green). However, it struggles to conduct further analysis to identify alignments in GMO related to the relationship between ``Award'' and ``hasCoPrincipalInvestigator'' in GBO. Rather, it attempts to generate random information about recently detected entities such as ``FundingAward'', which hypothetically arises from its next-word-prediction logic (highlighted in yellow). However, the remaining generated information, such as ``gmo\#Program'' (highlighted in red), is not relevant to this component in GBO. Ultimately, it states that \emph{There were no direct matches found for hasCoPrincipalInvestigator or isCoPrincipalInvestigatorOf in the GMO ontology.} 

We also attempted zero-shot prompting at this stage, combining both the GMO file and the second prompt into a single input to GPT-4. This approach resulted in increased latency and a confused response.

\begin{figure}[tb]
\centering
 \includegraphics[keepaspectratio, width=.8\textwidth, height=0.6\textheight]{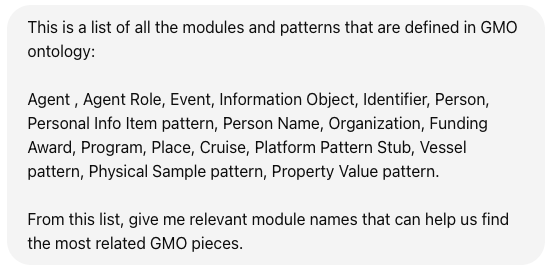}
 \caption{Prompt for module name suggestion} \label{m-name}
\end{figure}

Our conclusion from this analysis is that GPT-4 failed to detect the complex alignment between the two ontologies. Instead, it provided partially related objects, including random information about different unrelated entity classes.

\clearpage
To proceed, let us revisit the GMO modules discussed in Section \ref{method-1}. The documentation \cite{krisnadhi2015geolinkcorepattern} contains an informal description of the modules (in the documentation called \emph{patterns}) in the GMO ontology, accompanied by visual depictions of these patterns. In the ongoing chain-of-thoughts prompting processes, we included the module names and asked for the most relevant module names related to the GBO pieces, as shown in Figure \ref{m-name}. The result of this prompt is usually a single module or a list of suggested modules based on the information that the GPT has processed through the chain of prompts. In the next prompt, we included the descriptions of the suggested modules along with the question, as illustrated in the prompt below.
\begin{verbatim}
you couldn't give me all the pieces I need in GMO. here is more info that 
can help: The Funding Award pattern describes the funding awards that 
fund  all kinds of ocean science research activities. We use the 
isFundedBy property to connect anything to a funding award if the funding 
award funds it. Each funding award has exactly one starting and ending 
date (aligned with time:Instant). It provides at most one award amount, 
which is described via a pair of decimal value and currency code. The 
currency code is not specified here, but existing standards can be used, 
e.g., ISO  4217.  There may be people or organizations that have a role 
in a funding award. This is modeled by re-using (and aligning with) the 
Agent Role pattern. In this version, we include the following types of 
agent-roles, represented as classes: SponsorRole, 
AgencyProgramManagerRole, PrincipalInvestigatorRole, and 
CoPrincipalInvestigatorRole. Additional roles are possible in the future 
versions. Each funding award is described by an InformationObject, 
which when aligned to the Information Object pattern, allows one to 
represent additional information such as identifier, description, etc.

ObjectProperty: isFundedBy 
ObjectProperty: startsOnDate 
ObjectProperty: endsOnDate 
ObjectProperty: isDescribedBy
ObjectProperty: providesAgentRole 
ObjectProperty: isPerformedBy 
ObjectProperty: hasAwardAmount 
ObjectProperty: hasCurrencyCode 
DataProperty: hasCurrencyValue
Class: FundingAward
Class: Agent
Class: TimeInstant
Class: AgentRole
Class: SponsorRole
Class: PrincipalInvestigatorRole 
Class: CoPrincipalInvestigatorRole 
Class: AgencyProgramManagerRole 
Class: InformationObject
Class: AwardAmount 
Class: CurrencyCode
\end{verbatim}

\begin{figure}[tb]
\centering
 \includegraphics[width=\textwidth]{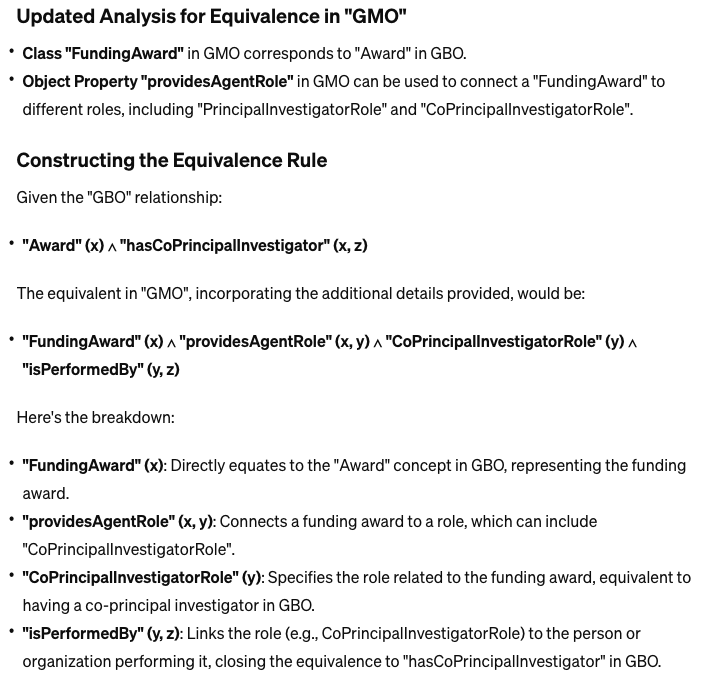}
 \caption{GPT-4 response to our question with Module information included} \label{fig-pr-4}
\end{figure}

By evaluating the outcome (see Fig. \ref{fig-pr-4}) of the prompt, it is evident that GPT-4 has a more informed and precise approach to conducting investigations and identifying semantically related components if it has the module information available. GPT-4 identifies all components related to the GMO as outlined in the alignment rule mentioned above. It further elaborates on each component and explains their interconnections. The full evaluation data of our study over all complex alignment rules is available online.\footnote{\url{https://github.com/reddraa/complex\_ontology\_alignment}}

\textbf{Impact of few-shot vs chain-of-thought:}
Comparing few-shot and chain-of-thought approaches, we noted the differences between providing information in a single zero-shot prompt versus delivering it in a series of prompts. GPT tends to become confused about the question and the relevant information it needs to process in a zero-shot scenario. In contrast, introducing the prompt as part of a sequential chain of information clarifies the data pieces and their meanings for the model.

\textbf{Impact of adding modules to the prompt:}
We observed that while GPT-4 nearly grasps the query regarding the relevant components needed to fulfill the rule, it typically identifies only 10-20\% of the required elements if no module information is given. However, if details about the module in the GMO are presented in a single prompt before posing the question again, GPT-4 is significantly more effective, clearly identifying the majority of the targeted components.

\subsection{Quantitative Evaluation}

To assess the effectiveness of our approach, we identified key entities within our complex alignment rules to serve as the basis for our metrics. As illustrated in Figure \ref{fig-pr-4}, the entities detected as relevant by GPT-4 include FundingAwards(z), providesAgentRole(x,y), CoPrinciplaInvestigatorRole(y), isPerformedBy(y,z). It is important to note that response formats may vary, hence each response might be unique. We employed two key performance metrics: recall and precision. Recall, defined as the ratio of detected GMO-related instances to the total number of expected instances, evaluates our study's ability to identify all relevant GMO pieces comprehensively. Precision, on the other hand, measures the accuracy of our detection process by calculating the proportion of correctly identified GMO pieces out of all the instances flagged in our findings. Together, these metrics provide a holistic view of our study's identification capability, striking a balance between thoroughness and accuracy in detecting GMO-related instances of a complex alignment rule. The well-known definitions are as follows.

\begin{align*}
\text{Recall} &= \frac{\text{Number of Correctly Detected Pieces}}{\text{Total Number of GMO Pieces in Complex Alignment}}\\
\text{Precision} &= \hspace*{12mm}\frac{\text{Number of Correctly Detected Pieces}}{\text{Total Number of Detected Pieces}}
\end{align*}

According to Table \ref{tbl-cov1}, in the study presented in the paper, an analysis of 109 complex alignment rules was conducted. The findings revealed that in only 4.5\% of these alignments, GMO components were identified solely through the use of GBO entities (i.e., not using module information). Both identified complex alignment pieces, without the module, are actually simple 1 $\colon$ 1 mappings, e.g., Program(x) $\leftrightarrow$ Program(x), where 'Program' is an existing class in both the GMO and GBO ontology. However, in over 95\% of the cases, the identification of GMO components was achieved through the application of information from the GMO module information, highlighting its significant role in the detection process.

\begin{table}[tb]
\centering
\caption{Distribution of successful approaches for detecting complex alignment rules}
\begin{tabular}{c|c}
\multicolumn{1}{c|}{detected pieces without module Information} & 5 \\ \hline
\multicolumn{1}{c|}{detected pieces with module Information} & 104 \\ \hline
\multicolumn{1}{c|}{total number of complex alignment rules} & 109 
\label{tbl-cov1}
\end{tabular}
\end{table}


In assessing GPT-4's performance on entity detection in the complex alignment, we analyzed the example alignment mentioned earlier in this paper.  We compared the alignment pieces returned by GPT-4, shown in Figure \ref{fig-pr-4} for our running example based on rule (\ref{eq:example-rule}), with four corresponding entities defined in the GMO ontology's alignment rule, i.e., FundingAward(x),  providesAgentRole(x,y),  CoPrincipalInvestigatorRole(y), performedBy(y,z). Note that we did not assess the actual return of alignment rules, but rather whether the relevant pieces (predicates) were detected. While the detection (without actual composition of the pieces into a rule) constitutes a simpler task than producing the rules, it nevertheless captures the core difficulty for complex ontology alignment. In fact, if the pieces are correct, the actual rule can easily be assembled by a human (or, in many cases, by a symbolic algorithm based on the ontology and example data). 

In evaluating recall (coverage) in this setting, we found that all four expected GMO entities were accurately identified by GPT-4 for our running example, yielding a recall of 1.0 in this case. Additionally, for precision, we examined the entities returned by GPT-4 and found that aside from the four correct GMO entities, no irrelevant entities were detected, also resulting in a precision of 1.0. This indicates perfect alignment detection by GPT-4 in this instance. To further clarify the evaluation of recall, consider that if the prompt response from GPT-4 had included additional entities such as ``Event'' or ``Place'' which were not part of the expected entities, it would have negatively impacted the precision. These extraneous entities, not being included in the set of expected results, would reduce the precision as they represent incorrect identifications according to the specified alignment rule. In practical terms, the return of (in particular, many) superfluous entities by the system would make assembly of the actual rule by a human more difficult.

In Table \ref{tbl-percentage} we see that for 73.3\% of the complex alignment rules evaluated, the recall value exceeded 0.5. This indicates that more than half of the GMO entities involved in a complex alignment rule were successfully detected for this ratio in our population. Furthermore, the recall value surpasses 0.75 for approximately 62.3\% of these rules, signifying a higher accuracy in detection. It's also noteworthy that \emph{for 45\% of the alignment rules recall is a perfect 1.0, and for 45.8\% precision is also 1.0}, because of the integration of module information in the analysis process.

The precision metric indicates the accuracy of the responses in directing us toward expected entities within the GMO ontology. Our analysis found that responses achieved a precision higher than 0.75 for 59.6\% of the evaluated records. Additionally, when the precision threshold is lowered to 0.5—implying that half of the entities suggested by the language model are the expected ones, while the other half may include relevant or irrelevant entities—the coverage of alignment rules increases to 69.7\%.

Note that detection (recall) of half or more of the correct body predicates, paired with high precision, is already very helpful for human assembly of a rule.

The recall data exhibits a mean of 0.67 and a median of 0.75, indicating that half of the recall values exceed 0.75, while the other half fall below this threshold. This distribution highlights that the majority of our data points demonstrate recall values above 0.75, offering insight into the central tendency of our dataset. Similarly, for precision, both the median and mean values suggest promising results, with at least 50\% of the records achieving a precision greater than 0.8. This underscores a generally reliable level of accuracy in the data.

\begin{table}[tb]
\centering
\caption{Recall and Precision for detected GMO entities using module information}
\label{tbl-percentage}
\begin{tabular}{c|p{1cm}|p{1cm}|p{1cm}|p{1cm}|p{1cm}|p{1cm}}
& \multicolumn{3}{c|}{Recall} & \multicolumn{3}{c}{Precision} \\ \cline{2-7} 
&\ $\geq  0.5$\ \ &\ $\geq 0.75$\ \  &\ $= 1$\ \  &\ $\geq 0.5$\ \  &\ $\geq  0.75$\ \  &\ $=  1$\ \  \\ \hline

\multicolumn{1}{c|}{with Module Information} &\ 73.3\%\ \ &\  62.3\%\ \  &\ 45.0\%\ \ &\ 69.7\%\ \ &\ 59.6\%\ \  &\ 45.8\%\ \  \\ \hline

\multicolumn{1}{c|}{mean} &\multicolumn{3}{c|}{0.67} & \multicolumn{3}{c}{0.67}  \\ \hline

\multicolumn{1}{c|}{median} &\multicolumn{3}{c|}{0.75} & \multicolumn{3}{c}{0.80}  \\ \hline

\multicolumn{1}{c|}{standard deviation} &\multicolumn{3}{c|}{0.37} & \multicolumn{3}{c}{0.37}  \\ \hline

\end{tabular}
\end{table}


We close the discussion with a number of additional more detailed observations.
\begin{itemize}
\item Type and Class Type alignments: An intriguing insight from the analysis of complex alignment rules is the difficulty encountered with Type or Class alignment. This process frequently falls short in identifying pertinent alignments. A case in point is the alignment where GeoFeatureType(x) in GBO transitions to rdfs:subClassOf(x, Place) in GMO. In such instances, the GMO ontology attempts to consolidate individual classes into types or subclasses. Nonetheless, the absence of this specific information in the GMO’s module data impedes the language model's (LLM) ability to accurately discern these sophisticated alignments, resulting in the provision of erroneous information.
\item Rich Modules, Improved Discovery: Another observation is that modules enriched with detailed descriptions, core axioms or alignments, as discussed in the referenced paper \cite{krisnadhi2015geolink}, depict significantly higher accuracy in discovery. Specifically, modules that offer comprehensive information facilitate better recall and precision rates. For example, the Cruise() entity, which involves approximately 18 complex alignment rules, demonstrates an impressive average recall rate of 0.93. This indicates that the depth of information provided directly influences the effectiveness of discovery processes.
\end{itemize}

 

%% file: 5-discussion.tex
\section{Discussion and Future Work}\label{sec:discussion}

The quantitative results we have just presented indicate that our setting---i.e., under inclusion of module information---produces high precision and recall values in many cases. Our results demonstrate a significant improvement compared to this baseline. While, absolutely speaking, the quantitative results are still moderate, we are in fact presenting the \emph{very first} reasonably working approach for generating complex alignments that does not require shared individuals; as noted, shared individuals are far from a realistic setting in practice. In the latest evaluation of the participating systems in the OAEI complex alignment track, which was in 2021, most systems failed to detect any m:n complex alignments.\footnote{https://oaei.ontologymatching.org/2021/results/complex/geolink/index.html} Only two systems procuded complex alignments,\footnote{https://oaei.ontologymatching.org/2021/results/complex/popgeolink/index.html} but they required instance data. As such our contribution shows a path forward towards complex alignment in realistic settings, a challenge that had so far eluded researchers.

For our approach to work, we provided the LLM with module information, following the previously presented arguments that (1) ontologies without additional internal structure or meaningful additional information are often too ambiguous for automated complex alignment tasks, (2) module identification during the ontology design process can easily be provided by the ontology modelers at least as part of the documentation (while doing so post-hoc, by others, requires major efforts), and (3) providing such module information as part of ontologies (and/or their documentations) would likely significantly decrease the effort and cost of many ontology engineering tasks, including complex alignment~\cite{shimizu2023modular}. As such, this is also a (repeated) call for improving ontology modeling methods to additionally provide module structure. 

While our results, as presented, are very encouraging, substantial future investigations will be required to cast them into an ontology alignment system that can work autonomously at high precision. Intermediate steps could constitute human-in-the-loop approaches where a human ontology engineer receives suggestions from an LLM, e.g., as to the relevant modules for a question, post-processes the LLM responses by manually checking the small number of suggestions, and feeding the correct suggestions back to the LLM for obtaining more complete responses. This is in line with the idea of an assisting system that limits the number of checks for the human, as opposed to the vast number of potential checks that would have to be done manually without such a limiting system.

In the future, we also intend to extend out approach to additional datasets featuring complex alignments for both evaluation and analytical objectives. Furthermore, we aim to explore alternative representations of modules to LLMs and evaluate the model's performance with these variations. Fine-tuning existing LLMs, but also the integration of additional symbolic data or algorithms, e.g. pertaining to logical axioms that come with well-designed ontologies, and also the integration of traditional simple alignment algorithms to further assist with the complex alignment task are all on our path forward.

%% file: references.bbl
\begin{thebibliography}{10}

\bibitem{achiam2023gpt}
Josh Achiam, Steven Adler, Sandhini Agarwal, Lama Ahmad, Ilge Akkaya, Florencia~Leoni Aleman, Diogo Almeida, Janko Altenschmidt, Sam Altman, Shyamal Anadkat, et~al.
\newblock Gpt-4 technical report.
\newblock {\em arXiv preprint arXiv:2303.08774}, 2023.

\bibitem{algergawy2018results}
Alsayed Algergawy, Michelle Cheatham, Daniel Faria, Alfio Ferrara, Irini Fundulaki, Ian Harrow, Sven Hertling, Ernesto Jim{\'e}nez-Ruiz, Naouel Karam, Abderrahmane Khiat, et~al.
\newblock Results of the ontology alignment evaluation initiative 2018.
\newblock In {\em 13th International Workshop on Ontology Matching co-located with the 17th ISWC (OM 2018)}, volume 2288, pages 76--116, 2018.

\bibitem{DBLP:conf/semweb/AlgergawyFFFHHJ19}
Alsayed Algergawy, Daniel Faria, Alfio Ferrara, Irini Fundulaki, Ian Harrow, Sven Hertling, Ernesto Jim{\'{e}}nez{-}Ruiz, Naouel Karam, Abderrahmane Khiat, Patrick Lambrix, Huanyu Li, Stefano Montanelli, Heiko Paulheim, Catia Pesquita, Tzanina Saveta, Pavel Shvaiko, Andrea Splendiani, {\'{E}}lodie Thi{\'{e}}blin, C{\'{a}}ssia Trojahn, Jana Vatascinov{\'{a}}, Ondrej Zamazal, and Lu~Zhou.
\newblock Results of the ontology alignment evaluation initiative 2019.
\newblock In Pavel Shvaiko, J{\'{e}}r{\^{o}}me Euzenat, Ernesto Jim{\'{e}}nez{-}Ruiz, Oktie Hassanzadeh, and C{\'{a}}ssia Trojahn, editors, {\em Proceedings of the 14th International Workshop on Ontology Matching co-located with the 18th International Semantic Web Conference {(ISWC} 2019), Auckland, New Zealand, October 26, 2019}, volume 2536 of {\em {CEUR} Workshop Proceedings}, pages 46--85. CEUR-WS.org, 2019.

\bibitem{amini2020geolink}
Reihaneh Amini, Lu~Zhou, and Pascal Hitzler.
\newblock Geolink cruises: A non-synthetic benchmark for co-reference resolution on knowledge graphs.
\newblock In {\em Proceedings of the 29th ACM International Conference on Information \& Knowledge Management}, pages 2959--2966, 2020.

\bibitem{brown2020language}
Tom Brown, Benjamin Mann, Nick Ryder, Melanie Subbiah, Jared~D Kaplan, Prafulla Dhariwal, Arvind Neelakantan, Pranav Shyam, Girish Sastry, Amanda Askell, et~al.
\newblock Language models are few-shot learners.
\newblock {\em Advances in neural information processing systems}, 33:1877--1901, 2020.

\bibitem{cheatham2014conference}
Michelle Cheatham and Pascal Hitzler.
\newblock Conference v2. 0: An uncertain version of the oaei conference benchmark.
\newblock In {\em The Semantic Web--ISWC 2014: 13th International Semantic Web Conference, Riva del Garda, Italy, October 19-23, 2014. Proceedings, Part II 13}, pages 33--48. Springer, 2014.

\bibitem{cheatham2018geolink}
Michelle Cheatham, Adila Krisnadhi, Reihaneh Amini, Pascal Hitzler, Krzysztof Janowicz, Adam Shepherd, Tom Narock, Matt Jones, and Peng Ji.
\newblock The geolink knowledge graph.
\newblock {\em Big Earth Data}, 2(2):131--143, 2018.

\bibitem{DBLP:books/daglib/0032976}
J{\'{e}}r{\^{o}}me Euzenat and Pavel Shvaiko.
\newblock {\em Ontology Matching, Second Edition}.
\newblock Springer, 2013.

\bibitem{DBLP:conf/semweb/0008C0023}
Yuan He, Jiaoyan Chen, Hang Dong, and Ian Horrocks.
\newblock Exploring large language models for ontology alignment.
\newblock In Irini Fundulaki, Kouji Kozaki, Daniel Garijo, and Jos{\'{e}}~Manu{\'{e}}l G{\'{o}}mez{-}P{\'{e}}rez, editors, {\em Proceedings of the {ISWC} 2023 Posters, Demos and Industry Tracks: From Novel Ideas to Industrial Practice co-located with 22nd International Semantic Web Conference {(ISWC} 2023), Athens, Greece, November 6-10, 2023}, volume 3632 of {\em {CEUR} Workshop Proceedings}. CEUR-WS.org, 2023.

\bibitem{hertling2023olala}
Sven Hertling and Heiko Paulheim.
\newblock Olala: Ontology matching with large language models.
\newblock In {\em Proceedings of the 12th Knowledge Capture Conference 2023}, pages 131--139, 2023.

\bibitem{DBLP:journals/cacm/Hitzler21}
Pascal Hitzler.
\newblock A review of the semantic web field.
\newblock {\em Commun. {ACM}}, 64(2):76--83, 2021.

\bibitem{FOST}
Pascal Hitzler, Markus Kr{\"{o}}tzsch, and Sebastian Rudolph.
\newblock {\em Foundations of Semantic Web Technologies}.
\newblock Chapman and Hall/CRC Press, 2010.

\bibitem{DBLP:series/faia/369}
Pascal Hitzler, Md.~Kamruzzaman Sarker, and Aaron Eberhart, editors.
\newblock {\em Compendium of Neurosymbolic Artificial Intelligence}, volume 369 of {\em Frontiers in Artificial Intelligence and Applications}.
\newblock {IOS} Press, 2023.

\bibitem{krisnadhi2015geolinkcorepattern}
Adila Krisnadhi, Yingjie Hu, Robert Arko, Suzanne Carbotte, Cynthia Chandler, Michelle Cheatham, Douglas Fils, Timothy Finin, Pascal Hitzler, Krzysztof Janowicz, et~al.
\newblock Geolink core ontology design patterns.
\newblock Technical report.
\newblock Available from \url{https://people.cs.ksu.edu/~hitzler/pub2/gmo-tr.pdf}.

\bibitem{krisnadhi2015geolink}
Adila Krisnadhi, Yingjie Hu, Krzysztof Janowicz, Pascal Hitzler, Robert Arko, Suzanne Carbotte, Cynthia Chandler, Michelle Cheatham, Douglas Fils, Timothy Finin, et~al.
\newblock The geolink modular oceanography ontology.
\newblock In {\em The Semantic Web-ISWC 2015: 14th International Semantic Web Conference, Bethlehem, PA, USA, October 11-15, 2015, Proceedings, Part II 14}, pages 301--309. Springer, 2015.

\bibitem{liu2023pre}
Pengfei Liu, Weizhe Yuan, Jinlan Fu, Zhengbao Jiang, Hiroaki Hayashi, and Graham Neubig.
\newblock Pre-train, prompt, and predict: A systematic survey of prompting methods in natural language processing.
\newblock {\em ACM Computing Surveys}, 55(9):1--35, 2023.

\bibitem{DBLP:conf/om2/NorouziMH23}
Sanaz~Saki Norouzi, Mohammad~Saeid Mahdavinejad, and Pascal Hitzler.
\newblock Conversational ontology alignment with chatgpt.
\newblock In Pavel Shvaiko, J{\'{e}}r{\^{o}}me Euzenat, Ernesto Jim{\'{e}}nez{-}Ruiz, Oktie Hassanzadeh, and C{\'{a}}ssia Trojahn, editors, {\em Proceedings of the 18th International Workshop on Ontology Matching co-located with the 22nd International Semantic Web Conference {(ISWC} 2023), Athens, Greece, November 7, 2023}, volume 3591 of {\em {CEUR} Workshop Proceedings}, pages 61--66. CEUR-WS.org, 2023.

\bibitem{DBLP:conf/semweb/PourAAAFFFHHHHI21}
Mina Abd~Nikooie Pour, Alsayed Algergawy, Florence Amardeilh, Reihaneh Amini, Omaima Fallatah, Daniel Faria, Irini Fundulaki, Ian Harrow, Sven Hertling, Pascal Hitzler, Martin Huschka, Liliana Ibanescu, Ernesto Jim{\'{e}}nez{-}Ruiz, Naouel Karam, Amir Laadhar, Patrick Lambrix, Huanyu Li, Ying Li, Franck Michel, Engy Nasr, Heiko Paulheim, Catia Pesquita, Jan Portisch, Catherine Roussey, Tzanina Saveta, Pavel Shvaiko, Andrea Splendiani, C{\'{a}}ssia Trojahn, Jana Vatascinov{\'{a}}, Beyza Yaman, Ondrej Zamazal, and Lu~Zhou.
\newblock Results of the ontology alignment evaluation initiative 2021.
\newblock In Pavel Shvaiko, J{\'{e}}r{\^{o}}me Euzenat, Ernesto Jim{\'{e}}nez{-}Ruiz, Oktie Hassanzadeh, and C{\'{a}}ssia Trojahn, editors, {\em Proceedings of the 16th International Workshop on Ontology Matching co-located with the 20th International Semantic Web Conference {(ISWC} 2021), Virtual conference, October 25, 2021}, volume 3063 of {\em {CEUR} Workshop Proceedings}, pages 62--108. CEUR-WS.org, 2021.

\bibitem{DBLP:conf/semweb/PourAAFFHHJJKKL20}
Mina Abd~Nikooie Pour, Alsayed Algergawy, Reihaneh Amini, Daniel Faria, Irini Fundulaki, Ian Harrow, Sven Hertling, Ernesto Jim{\'{e}}nez{-}Ruiz, Cl{\'{e}}ment Jonquet, Naouel Karam, Abderrahmane Khiat, Amir Laadhar, Patrick Lambrix, Huanyu Li, Ying Li, Pascal Hitzler, Heiko Paulheim, Catia Pesquita, Tzanina Saveta, Pavel Shvaiko, Andrea Splendiani, {\'{E}}lodie Thi{\'{e}}blin, C{\'{a}}ssia Trojahn, Jana Vatascinov{\'{a}}, Beyza Yaman, Ondrej Zamazal, and Lu~Zhou.
\newblock Results of the ontology alignment evaluation initiative 2020.
\newblock In Pavel Shvaiko, J{\'{e}}r{\^{o}}me Euzenat, Ernesto Jim{\'{e}}nez{-}Ruiz, Oktie Hassanzadeh, and C{\'{a}}ssia Trojahn, editors, {\em Proceedings of the 15th International Workshop on Ontology Matching co-located with the 19th International Semantic Web Conference {(ISWC} 2020), Virtual conference (originally planned to be in Athens, Greece), November 2, 2020}, volume 2788 of {\em {CEUR} Workshop Proceedings}, pages 92--138. CEUR-WS.org, 2020.

\bibitem{DBLP:conf/om2/PourABCCCC0FFFH23}
Mina Abd~Nikooie Pour, Alsayed Algergawy, Patrice Buche, Leyla~Jael Castro, Jiaoyan Chen, Adrien Coulet, Julien Cufi, Hang Dong, Omaima Fallatah, Daniel Faria, Irini Fundulaki, Sven Hertling, Yuan He, Ian Horrocks, Martin Huschka, Liliana Ibanescu, Sarika Jain, Ernesto Jim{\'{e}}nez{-}Ruiz, Naouel Karam, Patrick Lambrix, Huanyu Li, Ying Li, Pierre Monnin, Engy Nasr, Heiko Paulheim, Catia Pesquita, Tzanina Saveta, Pavel Shvaiko, Guilherme Sousa, C{\'{a}}ssia Trojahn, Jana Vatascinov{\'{a}}, Mingfang Wu, Beyza Yaman, Ondrej Zamazal, and Lu~Zhou.
\newblock Results of the ontology alignment evaluation initiative 2023.
\newblock In Pavel Shvaiko, J{\'{e}}r{\^{o}}me Euzenat, Ernesto Jim{\'{e}}nez{-}Ruiz, Oktie Hassanzadeh, and C{\'{a}}ssia Trojahn, editors, {\em Proceedings of the 18th International Workshop on Ontology Matching co-located with the 22nd International Semantic Web Conference {(ISWC} 2023), Athens, Greece, November 7, 2023}, volume 3591 of {\em {CEUR} Workshop Proceedings}, pages 97--139. CEUR-WS.org, 2023.

\bibitem{DBLP:journals/corr/abs-2312-00326}
Zhangcheng Qiang, Weiqing Wang, and Kerry Taylor.
\newblock Agent-om: Leveraging large language models for ontology matching.
\newblock {\em CoRR}, abs/2312.00326, 2023.

\bibitem{owl-primer}
Sebastian Rudolph, Markus Kr{\"{o}}tzsch, Peter Patel-Schneider, Pascal Hitzler, and Bijan Parsia.
\newblock {OWL} 2 web ontology language primer (second edition).
\newblock {W3C} recommendation, W3C, December 2012.
\newblock https://www.w3.org/TR/2012/REC-owl2-primer-20121211/.

\bibitem{shimizu2023modular}
Cogan Shimizu, Karl Hammar, and Pascal Hitzler.
\newblock Modular ontology modeling.
\newblock {\em Semantic Web}, 14(3):459--489, 2023.

\bibitem{shvaiko2011ontology}
Pavel Shvaiko and J{\'e}r{\^o}me Euzenat.
\newblock Ontology matching: state of the art and future challenges.
\newblock {\em IEEE Transactions on knowledge and data engineering}, 25(1):158--176, 2011.

\bibitem{wei2022chain}
Jason Wei, Xuezhi Wang, Dale Schuurmans, Maarten Bosma, Fei Xia, Ed~Chi, Quoc~V Le, Denny Zhou, et~al.
\newblock Chain-of-thought prompting elicits reasoning in large language models.
\newblock {\em Advances in Neural Information Processing Systems}, 35:24824--24837, 2022.

\bibitem{white2023prompt}
Jules White, Quchen Fu, Sam Hays, Michael Sandborn, Carlos Olea, Henry Gilbert, Ashraf Elnashar, Jesse Spencer-Smith, and Douglas~C Schmidt.
\newblock A prompt pattern catalog to enhance prompt engineering with chatgpt.
\newblock {\em arXiv preprint arXiv:2302.11382}, 2023.

\bibitem{zhou2018complex}
Lu~Zhou, Michelle Cheatham, Adila Krisnadhi, and Pascal Hitzler.
\newblock A complex alignment benchmark: Geolink dataset.
\newblock In {\em The Semantic Web--ISWC 2018: 17th International Semantic Web Conference, Monterey, CA, USA, October 8--12, 2018, Proceedings, Part II 17}, pages 273--288. Springer, 2018.

\bibitem{zhou2020geolink}
Lu~Zhou, Michelle Cheatham, Adila Krisnadhi, and Pascal Hitzler.
\newblock Geolink data set: A complex alignment benchmark from real-world ontology.
\newblock {\em Data Intelligence}, 2(3):353--378, 2020.

\end{thebibliography}
